\documentclass{llncs}
\usepackage{graphicx}
\usepackage{amsmath}
\usepackage{arydshln}

\usepackage{booktabs}
\usepackage{lipsum}
\usepackage{blindtext,graphicx}
\usepackage[absolute]{textpos}
\begin{document}

\begin{textblock}{10}(1,1)
\noindent\tiny This is a pre-print of an article published in MICCAI 2018.
\end{textblock}

\title{Magnetic Resonance Spectroscopy Quantification using Deep Learning}
\titlerunning{MRS Quantification with Deep Learning}  
%
\author{Nima Hatami \and Micha\"el Sdika \and H\'el\`ene Ratiney}
\authorrunning{Hatami et al.} 
%
%

\index{Hatami, Nima}
\index{Sdika, Micha\"el}
\index{Ratiney, H\'el\`ene}

\institute{Univ. Lyon, INSA-Lyon, Universit\'e Claude Bernard Lyon 1, UJM-Saint Etienne, CNRS, Inserm, CREATIS UMR 5220, U1206, F-69100, Lyon, France}



\maketitle              

\begin{abstract}
Magnetic resonance spectroscopy (MRS) is an important technique in biomedical research and it has the 
unique capability to give a non-invasive access to the biochemical content (metabolites) of scanned organs. 
In the literature, the quantification (the extraction of the potential biomarkers from the MRS 
signals) involves the resolution of an inverse problem based on a parametric model of 
the metabolite signal. However, poor signal-to-noise ratio (SNR), presence of the macromolecule signal or high correlation between metabolite spectral patterns can cause high uncertainties for most 
of the metabolites, which is one of the main reasons that prevents use of 
MRS in clinical routine. In this paper, quantification of metabolites in MR Spectroscopic imaging 
using deep learning is proposed. A regression framework based on the Convolutional Neural Networks (CNN) 
is introduced for an accurate estimation of spectral parameters. The proposed model learns the 
spectral features from a large-scale simulated data set with different variations of human brain 
spectra and SNRs. Experimental results demonstrate the accuracy of the proposed method, compared to 
state of the art standard quantification method (QUEST), on concentration of 20 metabolites and the macromolecule.    

\end{abstract}

\section{Introduction}

Magnetic Resonance Spectroscopy Imaging (MRSI) allows detection and localization of spectra from several spatially distributed voxels. After each voxel signal quantification, it  provides spatially resolved, non invasive and non-ionizing, metabolic information about the human body. The quantification process consists in analyzing the acquired spectra in order to estimate the metabolite concentrations, i.e. crucial biochemical information about the living cells and tissues. 

\subsection{MRS Quantification: problematic and state of the art}


MRS signals are acquired in the time domain, but are usually inspected in the frequency domain as the metabolites are characterized by specific spectral patterns. A salient aspect of MRS is that the concentration of one molecule is directly proportional to the signal amplitude in the resulting signal. The signals acquired with short echo time, which is the focus of this paper, contain several (up to 20) metabolite contributions and also a macromolecular background. The MRS signal $y(t)= x(t) + b(t) + e$ can be described as parametric (metabolites' part $x(t)$) and non-parametric parts. $x(t)$ is defined as a linear combination of metabolite signals. $b(t)$  is called the background signal: originating from macromolecules, it is qualified as non-parametric because its model function is not known (partially at least). In addition, acquisition artifacts (such as eddy current effect or water residual) and Gaussian random noise $e$ affect the acquired signal.  

Up to now, all the proposed quantification methods solve an optimization problem attempting  to minimize the difference between the data and a given parameterized model function. Most of the available methods employ local minimization and, in the case of short echo time, metabolite parameters are usually estimated by a non-linear least squares fit (in the time or the frequency domain) of the model (i.e. min $\lVert x -\hat{x} \rVert ^{2}$) using a known basis set of the metabolite signals . Despite numerous proposed fitting methods (for example QUEST\cite{quest}, LCModel\cite{Provencher1993lcmodel}, TARQUIN\cite{Wilson2011tarquin}), the robust, reliable and accurate quantification of brain metabolite concentration remains difficult. The major problems are: 
1) strong metabolite spectral pattern overlapping 
2) low signal to noise ratio, 
3) unknown background and peak line shape.
The problem is ill posed and current methods address it with different regularizations and constraint strategies 
(e.g. parameter bounds, penalizations), with possible large discrepancies in the results from one method to another \cite{Bhogal_2017}. 

Recently, as the application of machine learning expands into different domains, Das et. al \cite{das2017} applied the Random Forest regressor for MRS quantification. It creates a set of decision trees from randomly selected subset of training set. This is the first and so far the only machine learning approach applied to this problem. In their work, a simplified problem with only three to five metabolites is addressed. We compare their approach to ours in the experiments section.

\subsection{Contributions}

The contributions of the current work can be summarized as follows: i) addressing the MRS quantification problem using a deep learning approach for the first time. ii) proposing a synthetic MRS signal generation framework for the quantification purpose. 
Such a framework can not only simulate the \textit{in vivo} conditions, 
but also generate data free of cost and in a massive quantity. 
iii) proposing an appropriate CNN model that outperforms the state-of-the art fitting methods. iv) covering large number of metabolites (20) and the macromolecule. v) studying the effect of different noise levels.

The remainder of this paper is organized as follows: the next section gives an overview on MRS imaging, its quantification and the state-of the art fitting methods. The section 3 presents the proposed approach. The experiments, results and discussions are described in section 4. Section 5 concludes the paper and suggests the possible future directions.   

\section{MRS Quantification: A deep learning approach}

The mathematical model for the parametric part is defined as follows: 
\begin{eqnarray}
    x(t)=\sum_{m=1}^{M}a_{m}x_{m}(t) e^{\Delta \alpha_{m}t+2i\pi \Delta f_{m}t} 
    \label{signalModel}
\end{eqnarray}
where $M$ is the number of metabolites, $x_{m}(t)$ is the known ideal pattern of the \emph{m}th metabolite,
and the parameters to be estimated are the amplitude ($a_{m}$), the damping factor ($\Delta \alpha_{m}$) and the frequency shift ($\Delta f_{m}$). 
The amplitudes are directly proportional to the concentration of the metabolites.
The quantification process aims to find the parameters (amplitude, damping factor and frequency shift) for each metabolite in a way that the result fits the input signal. 


In this paper, a deep learning approach is presented as an alternative to the non-linear model fitting approaches of most methods in state of the art.
Instead of finding the signal parameters as the solution of an inverse problem between the partial model given by Eq. \ref{signalModel} and the signal,
our aim is to learn the inverse function once and for all on a training dataset.
Once this function is learnt, it can be used on a new signal for the quantification of its parameters.

The MRS quantification problem is converted from an online regression problem (robustly extracting the parameters by solving an inverse problem) to an
offline machine learning problem.
The process can be decomposed in three parts described in the paragraphs below:
one need to build the training dataset, to define a parametric representation of the inverse function 
and to setup a learning procedure to estimate the parameters of the inverse function.


\subsection{Data Generation Framework}

\begin{figure*}[t!]
\centering
\includegraphics[scale=0.3]{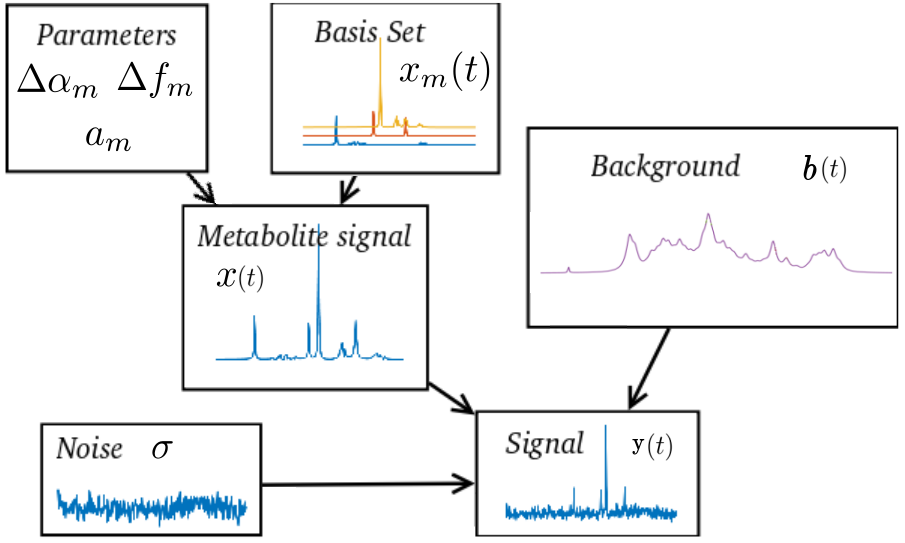}
\caption{The proposed synthetic MRS signal generation process.}
\label{figDataGeneration}
\end{figure*}

For any supervised learning technique to give satisfactory results, 
there should be enough training samples to be used in the learning process. 
Deep learning models in particular, require a relatively large amount of training data. 
A training dataset of \textit{in vivo} MRS signal cannot be built as it requires costly acquisition on human subject. Moreover ground truth metabolite concentrations are not available for in vivo signals, even by using medical experts.
%
%
This was the motivation to set up a synthetic data generation framework. 
The resulting dataset, if it succeeds to reproduce the distribution of realistic \textit{in vivo }signals,
has the advantage of being generated free of cost and on a massive scale.

The procedure to generate the dataset has been described in Fig. \ref{figDataGeneration}.
Metabolite parameters $a_m$, (resp. $\Delta \alpha_m$), (resp. $\Delta f_m$)
were randomly sampled with a distribution uniform in $[a_m^\text{min},a_m^\text{max}]$, 
(resp. $[-\Delta \alpha^\text{max},\Delta \alpha^\text{max}]$), 
(resp. $[-\Delta f^\text{max},\Delta f^\text{max}]$).
Knowing these parameters and the basis signals, the parametric signal $x$ can be computed using the equation \ref{signalModel}.
Here, the background was considered as another metabolite: random scaling factor, damping and frequency shift was
applied to the known background signal before it is added to $x$.
Random complex Gaussian noise is finally added to get the final signal.
To generate signal with a predefined SNR the standard deviation of the Gaussian distribution is set as the intensity of the first point of the noiseless signal
divided by the SNR.
This process can be repeated as many time as needed to create a large dataset of synthetic signals whose ground truth parameters are known.


\subsection{Convolutional Neural Networks}

There are two aspects of any CNN model that should be considered carefully: i) designing an appropriate architecture, and ii) choosing the right learning algorithm.  Both architecture and learning rules should be chosen in a way that they are not only compatible with each other, but also fit the data and the application
appropriately. 

\subsubsection{Architecture.}

CNN exploits spatially-local correlation by enforcing a local connectivity pattern between neurons of adjacent layers. Each layer is representing a different \emph{feature-level} and consists of convolution (filter), activation function, and pooling (a.k.a. subsampling), respectively. The input and output of each layer are called \emph{feature maps}. A filter layer convolves its input with a set of trainable kernels. The convolutional layer is the core building block of a CNN and exploits spatially local correlation by enforcing a local connectivity pattern between neurons of adjacent layers. The connections are local, but always extend along the entire depth of the input volume in order to produce the strongest response to a spatially local input pattern. 
Here we applied the recently proposed CReLU (Concatenated Rectified Linear Units) \cite{crelu} because it demonstrated improvement in the recognition performance. It is based on an observation in CNN models that the filters in lower layers form pairs (i.e. filters with opposite phase). To avoid the model to learn redundant filters of both positive and negative phase information, CReLU is proposed as follows:
\begin{eqnarray}
CRelu = Conc (r(x), -r(-x))
\end{eqnarray}
where, $Conc$ is the concatenation operator and ReLU is defined as $r(x)=max(0,x)$.

Pooling reduces the resolution of input and makes it robust to small variations for previously learned features. It combines the outputs of \emph{i-1}th layer into a single input in \emph{i}th layer over a range of local neighborhood. 

\begin{figure}[t!]
\includegraphics[scale=0.51]{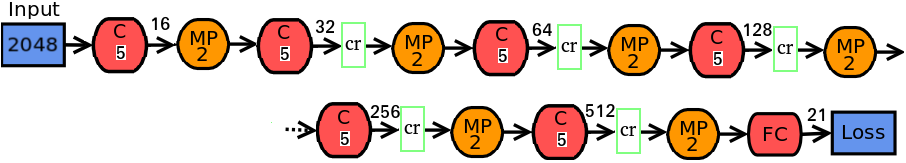}
\caption{The proposed CNN architecture for MRS quantification. The $C$, $MP$, $cr$, and $FC$ represent convolution, max-pool, CReLu, and fully-connected layers, respectively.}
\label{figCNN}
\end{figure}

At the end of the feature extraction layers, the feature maps are flatten and fed into a fully connected (FC) layer for regression. FC layers connect every neuron in one layer to every neuron in another layer, which in principle are the same as the traditional multi-layer perceptron (MLP). The proposed pipeline for MRS quantification is shown in Fig. \ref{figCNN}.

\subsubsection{Learning.}

Gradient-based optimization method (error back-propagation algorithm) is utilized to estimate parameters of the model. For faster convergence, the stochastic gradient descent (SGD) is used for updating the parameters. 
More details on CNN architecture and learning algorithm can be found in 
 \cite{lecun1998efficient,Bouvrie06}.

\section{Experiments and Results}

In the experiments, the metabolite basis set as well as the background signal provided by the ISMRM MRS Fitting Challenge 2016 
were used. Although all parameters were used to generate the signal, only the amplitude, which are the main parameters of interest,
were estimated by the neural network.
Amplitudes were drawn in $[0,1]$, $\Delta \alpha^\text{max}$ was set to 10Hz as well as $\Delta f^\text{max}$.

Training datasets of up to $5\times10^5$ samples were generated. 80\% of these samples are used to train the network and the rest is used 
as a validation dataset to evaluate the CNNs with different architectures, depths and solvers (optimization processes).
Once the best CNN is chosen, it is applied and compared to state of the art quantification methods on a different unseen test set
of 10,000 samples. 
As shown in Fig. \ref{figCNN}, a 7-layer CNN model is chosen with 2-channel (each real and imaginary part of the complex signal) input of size 2048 and the output layer with 21 neurons (20 metabolites amplitudes and a macromolecule scaling factor). 

The Symmetric mean absolute percentage error (SMAPE) \cite{SMAPE} over the whole test set is used to measure the accuracy of the models for each metabolite:
\begin{eqnarray}
SMAPE = \frac{\sum_{n=1}^{N} \mid a - \hat{a}\mid } {\sum_{n=1}^{N} (a + \hat{a})}
\end{eqnarray}
where $\hat{a}$ and $a$ are the estimated and ground truth amplitude values, respectively. 
SMAPE has been chosen as metric for its invariance to scale changes and its robustness to small values estimation.

Experiments were carried out using the Caffe framework \cite{caffe} with the Adam solver and the maximum number of iterations set to 200,000.
To initially move fast towards the local minimum, and move more slowly as approaching it, the "step" ($lr_0 \times \gamma^{floor(iter / step)})$ learning rate policy was chosen with $\gamma:0.5$ and $lr_0:10^{-3}$. 

\begin{figure}[t!]
\centering
\begin{minipage}{.48\textwidth}
  \centering
  \includegraphics[scale=0.45,keepaspectratio=true]{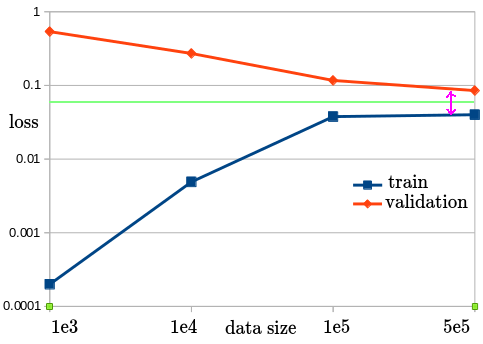}
  \caption{Learning curves: training and validation loss as a function of the training set size.
    The green line and pink gap approximately represent the estimated bias and standard deviation, respectively.}
  \label{figLC}
\end{minipage}%
\hspace*{\fill}
\begin{minipage}{.48\textwidth}
  \centering
  \includegraphics[scale=0.28,keepaspectratio=true]{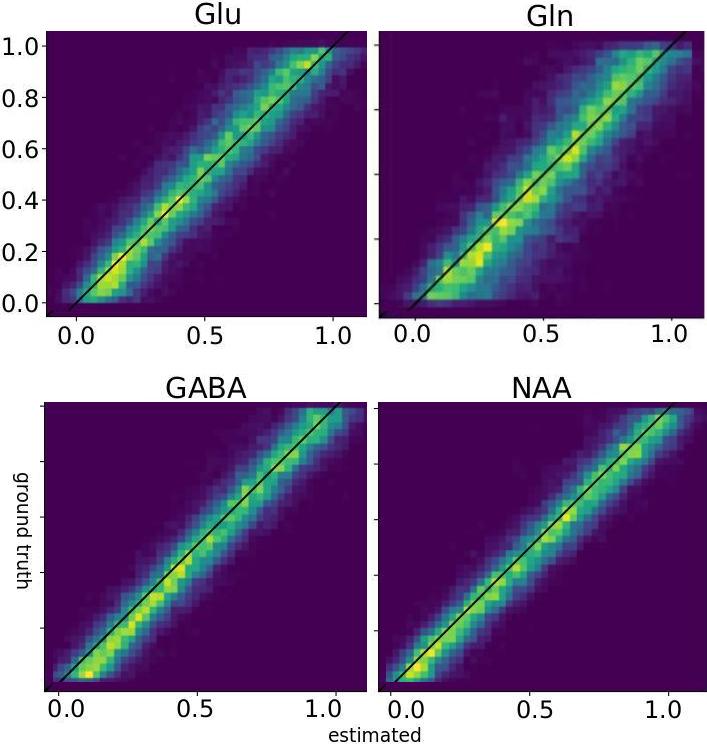}
  \caption{Ground truth vs. estimated metabolite concentrations using the CNN model for the test set (without noise).}
  \label{figXY}
\end{minipage}
\end{figure}

The less deep architectures have less parameters to adjust. 
Therefore, they need less data to train. 
However, for learning more complex tasks, 
expanding the layers is one of the options, which consequently requires larger data size.
In our case that the data can be generated in any desired size (except when there is time or computational restrictions),
we should try deeper models, if it was beneficial. The goal is to find an optimal architecture that minimizes the bias and standard deviation of the estimator. Fig. \ref{figLC} shows the process of choosing the optimal data size for a given CNN model.

For the comparison and as a gold standard, quantitation based on semi-parametric quantum estimation (QUEST) \cite{quest} is used. This nonlinear least-squares algorithm ranked between the best methods in the ISMRM'16 MRS Fitting Challenge. 
We also compare our results with the only machine learning approach applied on MRS quantification i.e. random forest regression algorithm \cite{das2017}. However, since the full details on the features used for the random forest is not given, we applied it on the raw data (no traditional hand-crafted feature extraction used).

\begin{table*}
\tiny
\caption{The SMAPE (\%) of the QUEST and Random Forest ensemble (RF) vs. the deep CNN model on the short echo-time data with no noise (left) and 10 SNR (right).}

\begin{center}
\begin{tabular}{lll}
\begin{tabular}{c c c c}
\hline
Metabolite & Quest & RandForest & CNN CReLu\\[2pt]
\hline\rule{0pt}{12pt}
Ala	 &       6.64      & 22.05 	& \bf 2.80 	\\
Asc      &       6.44      & 22.13 	& \bf 3.92 	\\
Asp      &       8.81      & 24.12 	& \bf 5.23 	\\
Cr       & \bf   10.31     & 20.70 	&     12.28 	\\
GABA     &       15.48     & 16.86 	& \bf 5.98 	\\
GPC      &       5.53      & 14.96 	& \bf 3.34 	\\
GSH      &       7.94      & 21.52 	& \bf 4.44 	\\
Glc      &       10.89     & 22.00 	& \bf 2.01 	\\
Gln      &       18.11     & 23.60 	& \bf 9.89 	\\
Glu      &       15.97     & 23.07 	& \bf 7.74 	\\
Gly      &       12.44     & 23.57 	& \bf 9.81 	\\
Ins      &       11.84     & 20.89 	& \bf 8.72 	\\
Lac      &       6.34      & 20.46 	& \bf 2.43 	\\
NAA      &       9.26      & 20.87 	& \bf 5.38 	\\
NAAG     &       7.15      & 15.75 	& \bf 3.76 	\\
PCho     &       6.13      & 16.10 	& \bf 4.94 	\\
PCr      &  \bf  10.24     & 20.67 	&     11.19 	\\
PE       &       17.64     & 24.26 	& \bf 10.96 	\\
Tau      &       14.81     & 23.25 	& \bf 11.65 	\\
sIns     &       6.80      & 16.91 	& \bf 6.10 	\\
\hdashline   
Macromol.&       1.32      & 5.06 	& \bf 0.86 	\\
[2pt]
\hline
\# wins   & 2              & 0          & \bf 19 \\
Ave. Rank & 2.47           & 2.52       &  \bf 1.00 \\ 
\hline
\end{tabular}
&
\hspace{1mm}
&
\begin{tabular}{c c c c}
\hline
Metabolite & Quest & RandForest  & CNN CReLu\\[2pt]
\hline\rule{0pt}{12pt}
Ala	 & 31.21  	& 24.39 	&     21.03 	\\
Asc      & 28.80  	& 24.24 	& \bf 20.64 	\\
Asp      & 41.98  	& \bf 25.38 	&     25.87 	\\
Cr       & 26.30  	& 22.92 	& \bf 19.64 	\\
GABA     & 37.63  	& 25.14 	& \bf 24.37 	\\
GPC      & 25.06  	& 19.33 	& \bf 13.67 	\\
GSH      & 26.81  	& 23.70 	& \bf 19.06 	\\
Glc      & 33.42  	& 24.07 	& \bf 20.85 	\\
Gln      & 36.09  	& 24.94 	& \bf 22.98 	\\
Glu      & 34.88  	& 24.82 	& \bf 22.29 	\\
Gly      & 29.78  	& 24.18 	& \bf 21.50 	\\
Ins      & 28.03  	& 23.73 	& \bf 20.20 	\\
Lac      & 28.80  	& 24.32 	& \bf 20.40 	\\
NAA      & 26.69  	& 22.93 	& \bf 18.95 	\\
NAAG     & 23.58  	& 21.85 	& \bf 16.03 	\\
PCho     & 21.44  	& 19.60 	& \bf 14.27 	\\
PCr      & 26.40  	& 22.77 	& \bf 19.39 	\\
PE       & 43.29  	& 24.84 	& \bf 23.29 	\\
Tau      & 36.82  	& 24.20 	& \bf 22.18 	\\
sIns     & 23.02  	& 17.72 	& \bf 14.09 	\\
\hdashline      
Macromol.& 14.45  	& 17.59 	& \bf 8.84 	\\[2pt]
\hline
\# wins   & 0           &  1            & \bf 20 \\
Ave. Rank & 2.95        & 2.00          & \bf 1.04 \\ 

\hline
\end{tabular}
\end{tabular}
\end{center}
\end{table*}


\subsubsection{Discussion:}
This work has  tackled, through the proposed deep learning approach, the major bottleneck of MRS quantification which is the metabolite peak overlapping and macromolecular background contamination.
One can also see that learning curves presented in Fig. \ref{figLC} has the expected shape: this will allow to 
estimate the bias and generalization power of our CNN estimator.
Remarkably, the SMAPE is high in QUEST for the metabolites that are known to have overlapped spectral pattern (and thus strong amplitude parameter correlation) such as GABA, Glu, Gln , but also Glc, Ins, sIns,  while CNN CRelu and RF performance appear to be insensitive to spectral pattern overlapping. This results can be confirmed visually on the plot presented in Fig. \ref{figXY}.
%
%
Results from Table 1 show that CNN quantification outperforms the two other methods both with and without noise.
One can notice that without noise, the QUEST 's SMAPE were smaller than RF while it is not the case for noisy data. Note that the chosen noise level is really important here and most of the acquisitions are generally done with higher SNRs. The obtained results demonstrate the high noise robustness of machine learning approaches. Finally, these different methods were compared on data which metabolite relative concentrations/proportions do not mimic \textit{in vivo} conditions. However, the present results demonstrate the ability of CNN to perform MRS quantification without being hampered by the usual limitations. 
The next step is to integrate more realistic signal in the data generation, for example by including phase variation due to eddy current, residual water or non ideal lineshapes.

\section{Conclusions and Future Work}
Quantification of metabolites in MRS imaging using deep learning is presented for the first time. A CNN model, as a class of deep, feed-forward artificial neural networks is used for accurate estimation of spectral parameters. Since efficient training of the CNN model requires large number of samples and such a data is not available  \textit{in vivo}, a new framework of generating a simulated human brain spectra is set up. Experiments are carried out on 20 metabolites and the macromolecule using different noise levels. The obtained results are compared to the Quest and the Random Forest regressor, highlighting the superiority of the proposed method. This study opens a new line of research to further investigate the application of deep learning techniques on MRS quantification problem.

Some future directions to extend the current work are i) validation of the proposed CNN model on\textit{ in vivo} data, ii) including the non-linear effects and artifacts (e.g. water residue and eddy current effect) in the synthetic data generation model for more realistic simulation of\textit{ in vivo} conditions, and iii) investigating different deep learning models
, architectures, and signal representations (e.g. image representation of spectral data \cite{hatami2017}) for improving the accuracy.

\section*{Acknowledgement}

This work is supported by the academic program of NVIDIA, the CNRS PEPS "APOCS" and the LABEX PRIMES (ANR-11-LABX-0063) of Universit\'e de Lyon, within the program "Investissements d'Avenir" (ANR-11-IDEX-0007).
We also acknowledge the CC-IN2P3 for providing the computing resources.



%
%

\end{document}